\PassOptionsToPackage{table}{xcolor}  % 放在最前面

\documentclass[pdflatex,sn-mathphys-num]{sn-jnl}% Math and Physical Sciences Numbered Reference Style
\usepackage{graphicx}%
\usepackage{multirow}%
\usepackage{amsmath,amssymb,amsfonts}%
\usepackage{amsthm}%
\usepackage{mathrsfs}%
\usepackage[title]{appendix}%
\usepackage{xcolor}%
\usepackage{textcomp}%
\usepackage{manyfoot}%
\usepackage{booktabs}%
\usepackage{algorithm}%
\usepackage{algorithmicx}%
\usepackage{algpseudocode}%
\usepackage{listings}%
\usepackage{xcolor}
\usepackage{bm}
\usepackage{caption}
\usepackage{makecell}
\theoremstyle{thmstyleone}%
\theoremstyle{thmstyletwo}%

\theoremstyle{thmstylethree}%

\begin{document}

\title[Article Title]{SynBridge: Bridging Reaction States via Discrete Flow for Bidirectional Reaction Prediction}


\author[1]{\fnm{Haitao} \sur{Lin}}\email{linht@aisi.ac.cn}\email{lbirdtao@gmail.com}

\author[2,3]{\fnm{Junjie} \sur{Wang}}\email{1800011822@pku.edu.cn}

\author[2]{\fnm{Zhifeng} \sur{Gao}}\email{gaozf@dp.tech}

\author[2]{\fnm{Xiaohong} \sur{Ji}}\email{jixh@dp.tech}

\author[1,3]{\fnm{Rong} \sur{Zhu}}\email{rongzhu@pku.edu.cn}

\author[2]{\fnm{Linfeng} \sur{Zhang}}\email{zhanglf@dp.tech}

\author[*,2]{\fnm{Guolin} \sur{Ke}}\email{kegl@dp.tech}

\author[*,1,4,5]{\fnm{Weinan} \sur{E}}\email{weinan@math.pku.edu.cn}

\affil[1]{\small \orgdiv{AI for Science Institute}, \orgaddress{\city{Bejing}, \postcode{100080}, \state{China}}}

\affil[2]{\small \orgdiv{DP Technology}, \orgaddress{\city{Bejing}, \postcode{100080}, \state{China}}}
\affil[3]{\small \orgdiv{College of Chemistry and Molecular Engineering}, \orgname{Peking University}}
\affil[4]{\small \orgdiv{School of Mathematical Sciences}, \orgname{Peking University}, \orgaddress{\city{Bejing}, \postcode{100871}, \state{China}}}
\affil[5]{\small \orgdiv{Center for Machine Learning Research}, \orgname{Peking University}, \orgaddress{\city{Bejing}, \postcode{100084}, \state{China}}}
\affil[*]{\small Corresponding Authors}

\abstract{The essence of a chemical reaction lies in the redistribution and reorganization of electrons, which is often manifested through electron transfer or the migration of electron pairs. These changes are inherently discrete and abrupt in the physical world, such as alterations in the charge states of atoms or the formation and breaking of chemical bonds. To model the transition of states, we propose SynBridge, a bidirectional flow-based generative model to achieve multi-task reaction prediction. By leveraging a graph-to-graph transformer network architecture and discrete flow bridges between any two discrete distributions, SynBridge captures bidirectional chemical transformations between graphs of reactants and products through the bonds' and atoms' discrete states. We further demonstrate the effectiveness of our method through extensive experiments on three benchmark datasets (USPTO-50K, USPTO-MIT, Pistachio), achieving state-of-the-art performance in both forward and retrosynthesis tasks. Our ablation studies and noise scheduling analysis reveal the benefits of structured diffusion over discrete spaces for reaction prediction.}




\maketitle

\section{Introduction}\label{sec1}

The ability to accurately predict chemical reaction outcomes, including both forward synthesis and retrosynthetic pathways, is a cornerstone task in computer-aided synthesis planning in the field of computational chemistry. It holds transformative implications for drug design, materials development, and green chemistry by reducing trial-and-error in synthetic route discovery~\cite{segler2018planning}.

Early computational efforts typically relied on rule-based expert systems or reaction templates manually encoded by chemists~\cite{corey1989retrosynthetic,law1986expert}. While interpretable, these methods suffer from poor scalability and generalization, being limited to previously seen transformations. In response, deep learning models have emerged as powerful alternatives, shifting from predefined rules to data-driven paradigms that learn reaction patterns from large-scale reaction datasets~\cite{schwaller2019molecular,coley2017prediction, jin2017predicting}.

Recent advances in large language models (LLMs) and generative modeling have further propelled the progress in chemical reaction modeling~\cite{devlin2018bert,brown2020language}. SMILES-based sequence-to-sequence models treat molecules as token sequences and employ LLM architectures to predict chemical products~\cite{schwaller2021mapping}. Although effective, these models often fail to capture molecular topology and spatial invariance. Graph-based approaches mitigate these limitations by directly modeling molecular structures as graphs, enabling more chemically grounded representations~\cite{coley2019graph,bi2021nerf,igashov2024retrobridge}. Nonetheless, many such models rely on autoregressive decoding, introducing inefficiencies through fixed generation orders and compounding errors during sequential prediction. Moreover, most existing one-shot graph-based models are task-specific, addressing only forward prediction or retrosynthesis individually~\cite{somnath2021learning}. In contrast, LLMs inherently support multitask capabilities, as exemplified by models such as {T5Chem}~\cite{kim2021t5chem}, which can simultaneously perform product generation, reactant prediction, and reagent suggestion. Bridging the structural precision of graph-based models with the task flexibility of LLMs remains a key open challenge in building unified frameworks for chemical reaction prediction.

To this end, we introduce {SynBridge}, a unified bidirectional framework for chemical reaction modeling, grounded in discrete diffusion principles. Our approach builds on the recent success of diffusion models in image and text generation~\cite{ho2020denoising,song2021ddim,song2021scorebased,rombach2022high,nichol2022glide,saharia2023imagen,Ma2024SiTECCV,gong2022diffuseq}, particularly the discrete variants suited for structured domains~\cite{liu2022flow}. {SynBridge} is built upon a discrete flow matching~\cite{gat2024discreteflowmatching} and diffusion bridge model~\cite{Albergo2023StochasticInterpolants,zhou2023ddbms,kim2025ddsbm}, tailored specifically for discreteness of molecular graphs. It iteratively transforms a source molecular graph into a target graph via learned probabilistic velocity fields. The forward process of the flow bridge corresponds to product prediction from reactants, while the reverse direction naturally aligns with retrosynthesis, generating plausible reactants from a given product. This bidirectional formulation enables {SynBridge} to unify both tasks within a single framework, offering a principled and efficient approach for multitask reaction prediction over discrete molecular graph spaces.
,

\section{Model}\label{sec:model}
\begin{figure}
    \centering
    \includegraphics[width=1.0\textwidth, trim=160 200 160 150, clip]{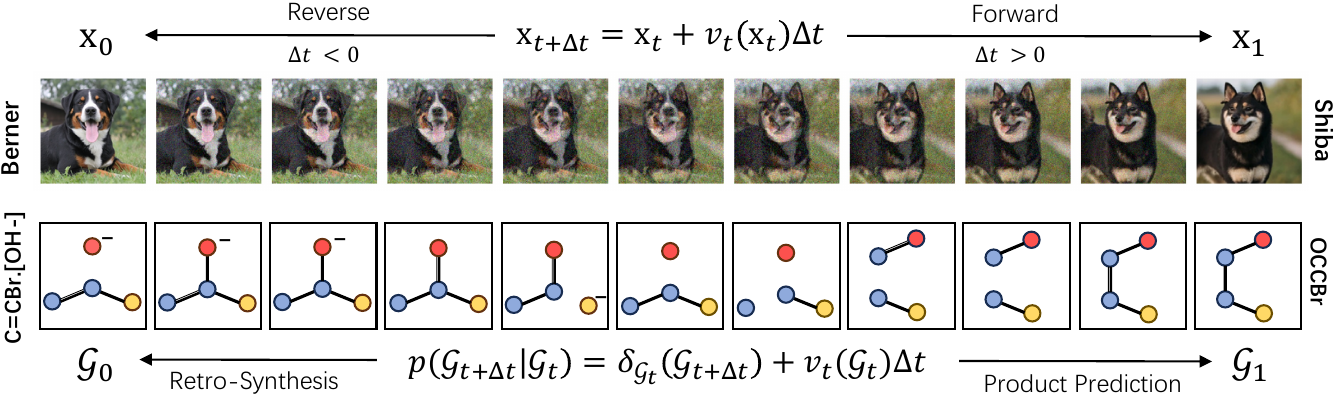}
    \caption{A schematic analogy: The image-to-image bridge models parameterize $v_t(\cdot)$ as the velocity field to translate between continuous image signals $\mathrm{x}_0$ and $\mathrm{x}_1$. In comparison, SynBridge regards graphs of molecules as discrete variables and translates between the reactant graphs $\mathcal{G}_0$ and the product graphs  $\mathcal{G}_1$ with discrete flow bridges. An Example of `\colorbox{cyan!20}{\texttt{C=C}}\colorbox{yellow!40}{\texttt{Br}}.\colorbox{red!20}{\texttt{[OH-]}}\texttt{>>} \colorbox{red!20}{\texttt{O}}\colorbox{cyan!20}{\texttt{CC}}\colorbox{yellow!40}{\texttt{Br}}' is given, while the intermediate graphs are sampled from the probability paths estimated by the bridge models.  }~\label{fig:analogy}
\end{figure}
SynBridge is built upon a Discrete Flow Bridge, which formulates the transformation from initial reactant graphs to the final product graphs and vice versa as graph-to-graph translation problems (Sec.~\ref{sec:taskformu}). Motivated by the recent advances of Diffusion Bridge Models in image-to-image translation, SynBridge leverages a discrete probability flow as a bridge model to translate between two arbitrary discrete distributions (Sec.~\ref{sec:discretefb}). In the forward flow path, it progressively pushes the reactant graph toward the product graph, as the task of forward product prediction, and conversely, the reverse process of the flow bridge naturally models backward transitions, enabling flexible retrosynthesis prediction in the reverse directions (Sec.~\ref{sec:learning}). Fig.~\ref{fig:analogy} presents a schematic analogy that highlights the conceptual parallel between image-to-image bridge models and SynBridge.

\subsection{Task Formulation}\label{sec:taskformu}
A natural way to express a reaction is to regard the combination of reactant molecules as a graph in which each molecule is one of the subgraphs in it, and likewise, the product molecule can be represented as a graph in which every atom has a unique correspondence to an atom in the reactant graph via a one-to-one atom mapping. Accordingly, product prediction in a chemical reaction can be formulated as a graph-to-graph translation task from the reactants, whereas reactant prediction, which lies at the core of retrosynthesis, corresponds to the inverse translation process. 

Formally, we define the reactant graph as the initial graph of a reaction consisting $N$ atoms, denoted by $\mathcal{G}_0 = (\mathcal{V}_0, \mathcal{E}_0)$, where $\mathcal{V}_0$ is the atom node sets and $\mathcal{E}_0$ is the bond sets, and similarly,  $\mathcal{G}_1 = (\mathcal{V}_1, \mathcal{E}_1)$ is the product graph as the final graph of the reaction. In the real-world scenario of retrosynthesis, usually, only the main product is available, leading to missing atoms that were present in the reactants. In this case, we treat those atoms that appear in the reactants but are absent in the main product as dummy atoms, and in the retrosynthesis task, one of the goals is to recover their correct element types as they exist in the reactants. Therefore, we consider the change of three atom node features, atom types, electron transfer, and aromaticity. In this way, the atom type variable $a^{(i)}$, binary aromaticity variable $b^{(i)}$ indicating whether the atom is aromatic, and charge variable $c^{(i)}$ are used to featurize the atom nodes, leading to $\mathcal{V} = \{(a^{(i)},b^{(i)}, c^{(i)})| 1\leq i\leq N\}$. Suppose that there are $M-1$ types of atom elements, so $a^{(i)} \in \{1, \ldots, M\}$, $b^{(i)}\in \{0,1\}$, and $c^{(i)} \in \{-6, -5, \ldots, 5, 6\}$ are discrete variables with $M$ as a `\texttt{dummy type}', $a^{(i)} \leq M-1$ if $i \in \text{main product}$ and $a^{(i)} = M$ if $i \in \text{reactants}\setminus \text{main product}$.
In addition, changes in bonds are also the key to expressing the reaction. We consider the adjacency matrix $\bm{R}$ determined by the bond set $\mathcal{E}$, elements in which are denoted by $r^{(i,j)} \in \{0,1,2,3\}$, indicating the bond types of `\texttt{nonbonded}', `\texttt{single}', `\texttt{double}' and `\texttt{triple}'. Note that the `aromatic bond' is classified as `single' since the atom's aromatic variable featurize the molecular graph. Since the four variables are all taking values in a discrete space, the translation of them needs a discrete probabilistic model instead of a commonly used continuous one. Fig.~\ref{fig:workflow}.a and c give an illustration of the descriptive variables for the reactants and products.

\subsection{Discrete Flow Bridge }\label{sec:discretefb}
\subsubsection{Probability Path and Velocity }

To start with, we firstly introduce a general flow bridge model that transforms source samples $\bm{x}_0 \sim p_0$ to target samples $\bm{x}_0 \sim p_1$, in which $p_0$ and $p_1$ are all discrete probability functions with $K$ classes, and $\bm{x}_0, \bm{x}_1 \in \{1,\ldots,K\}$.  To enable translation between discrete probabilities, we follow flow matching approaches~\cite{Lipman2023FlowMatchingICLR,Lipman2024FlowMatchingGuide,Albergo2024CFM,Ma2024SiTECCV} and establish a \textit{discrete flow bridge} model. Given a source-target pair \((\bm{x}_0, \bm{x}_1)\), we define their joint distribution as \(\pi(\bm{x}_0, \bm{x}_1)\). The associated probability path is defined as:
\begin{equation}
p_t(\bm{x}) = \sum_{\bm{x}_0, \bm{x}_1} p_t(\bm{x} | \bm{x}_0, \bm{x}_1)\, \pi(\bm{x}_0, \bm{x}_1),
\end{equation}
where \(p_t(\bm{x} | \bm{x}_0, \bm{x}_1)\) is referred to as the \textit{conditional probability path}. Inspired by stochastic interpolation over continuous variables~\cite{Albergo2023StochasticInterpolants}, which bridges two endpoint distributions via time-dependent Gaussian noise, we predefine the conditional path as:
\begin{equation}
p_t(\bm{x} | \bm{x}_0, \bm{x}_1) = \alpha_t\, \delta_{\bm{x}_0}(\bm{x}) + \beta_t\, \delta_{\bm{x}_1}(\bm{x}) + \sigma_t\, U_{\frac{1}{K}}(\bm{x}), \label{eq:cpp}
\end{equation}
where \(U_{\frac{1}{K}}(\cdot)\) denotes the uniform discrete distribution over all classes with equal probability \(1/K\), and \(\delta_{\bm{x}_i}(\cdot)\) represents the Kronecker delta function, assigning probability 1 when \(\bm{x} = \bm{x}_i\) and 0 otherwise. $\alpha_t$, $\beta_t$ and $\sigma_t$ are the coefficient that governs the changing rate of the conditional probability path, such that $\alpha_t + \beta_t+\sigma_t = 1$, $\alpha_1=\beta_0=0$, $\alpha_0 = \beta_1=1$, $\sigma_t\geq0$, $\sigma_0=\sigma_1=0$ are the uniform noise level. The inclusion of the uniform term is critical to improving the self-correction capability of the model by preventing early overcommitment to either endpoint distribution (see Sec.~\ref{sec:ablation} for details).
\begin{figure}
    \centering
    \includegraphics[width=1.0\textwidth, trim=82 100 90 65, clip]{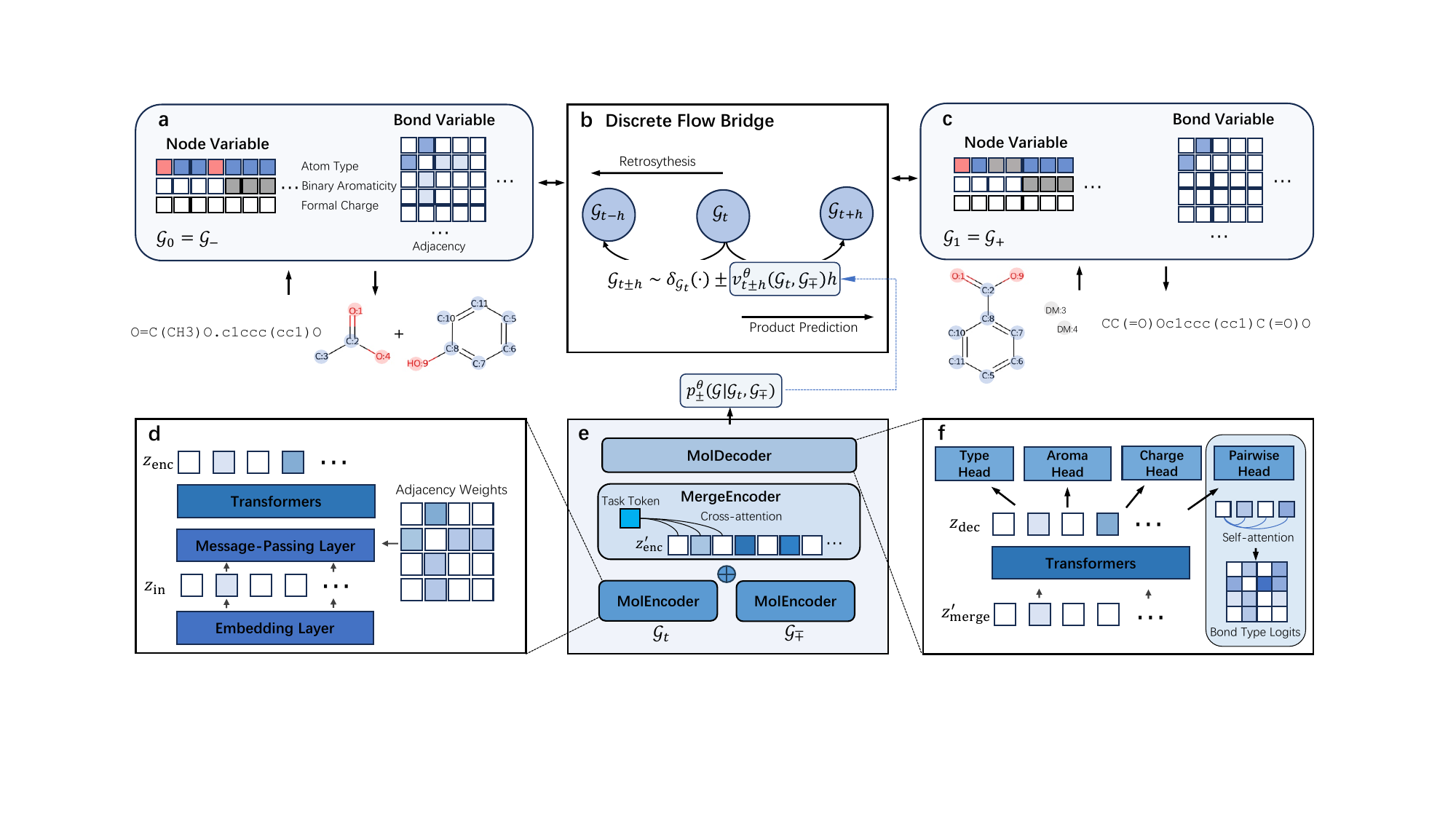}
    \caption{The workflow of SynBridge. \textbf{a}: The reactants are described by node variables including `atom type', `binary aromaticity' ,and `formal charge', and bond variables as the elements in the adjacency matrix. \textbf{b}: The \textit{Discrete Flow Bridge} model establishes a bidirectional reaction prediction bridge based on the learned probability velocity. \textbf{c}: In the node variables generated by products, atoms labeled as \colorbox{gray!12}{\texttt{DM}} represent `\texttt{Dummy types}', which are absent in the main product but in the reactants. \textbf{d}: The \textit{MolEncoder} architecture includes one node embedding layer with atom attributes as input, one single message-passing layer regulated by weights corresponding to elements in the adjacency matrix, followed by stacks of transformers for further feature extraction. \textbf{e}: The overall network architecture takes as input the current graph $\mathcal{G}_t$ and the source graph $\mathcal{G}_{\mp}$. After the encoder's extraction of $\bm{\mathrm{z}}_\mathrm{enc}$, \textit{MergeEncoder} performs cross-attention of task token over the atom-level molecular representations, leading to $\bm{\mathrm{z}}'_\mathrm{enc}$, which is then mapped by the decoder to the probability distribution over the target graph  $p_\pm^\theta(\mathcal{G}|\mathcal{G}_t, \mathcal{G}_\mp)$. \textbf{f}: The \textit{MolDecoder} consists of stacked transformer layers, transforming $\bm{\mathrm{z}}'_\mathrm{merge}$ to  $\bm{\mathrm{z}}_\mathrm{dec}$ . The final $\bm{\mathrm{z}}_\mathrm{dec}$ is passed through three linear heads to generate logits of probabilities of atom type, binary aromaticity, and formal charges, and a self-attention operation where the resulting attention matrix produces the logits of bond type probabilities.}~\label{fig:workflow}
\end{figure}

And thus, followed by \textbf{Theorem3} in \textit{Discrete Flow Matching} \cite{gat2024discreteflowmatching}, the generating conditional velocity based on Eq.~\ref{eq:cpp} for the probability path conditioned on last sample $\bm{x}_s$ with $s< t$  can be written as  
\begin{equation}
    v_t(\bm{x}| \bm{x}_s, \bm{x}_0, \bm{x}_1) = (\dot{\alpha}_t -\alpha_t\gamma_t) \delta_{\bm{x}_0}(\bm{x}) + (\dot{\beta}_t -\beta_t\gamma_t) \delta_{\bm{x}_1}(\bm{x}) + (\dot{\sigma}_t - \sigma_t\gamma_t)\, U_{\frac{1}{K}}(\bm{x}) + \gamma_t\delta_{\bm{x}_s}(\bm{x}), \label{eq:fwdvelocity}
\end{equation}
where $\gamma_t = \min\{\frac{\dot{\alpha}_t}{{\alpha}_t}, \frac{\dot{\beta}_t}{\beta_t}, \frac{\dot{\sigma}_t}{\sigma_t}\}$.

\subsubsection{Parameterization and Learning Objective}
To learn the velocity field $v_t(\bm{x}|\bm{x}_s, \bm{x}_0 , \bm{x}_1)$ from $t=0$ to $t=1$, a neural network $p_1^{\theta}(\cdot)$ with parameter $\theta$ can be used to approximate $p_1(\bm{x}|\bm{x_0}) = \delta_{\bm{x}_0}(\bm{x})$, with minimization of cross-entropy ($\mathrm{CE}$) loss as the learning objective
\begin{equation}
\begin{aligned}
        \theta^* &= \arg\min_\theta \mathcal{L}(\theta) \\&= \arg\min_\theta\mathbb{E}_{\substack{
    (\bm{x}_0, \bm{x}_1) \sim \pi(\bm{x}_0, \bm{x}_1) \\
    \bm{x}_s \sim p_s(\bm{x} \mid \bm{x}_0, \bm{x}_1)
    } } \mathrm{CE}\left(p_1^{\theta}(\bm{x}|\bm{x}_s,\bm{x}_0), \delta_{\bm{x}_1}(\bm{x}) \right) 
\end{aligned}
\end{equation}
leading to the parameterized conditional velocity of 
\begin{equation}
     v^{\theta}_t(\bm{x}|\bm{x}_s, \bm{x}_0) = (\dot{\alpha}_t -\alpha_t\gamma_t)  \delta_{\bm{x}_0}(\bm{x}) + (\dot{\beta}_t -\beta_t\gamma_t)p_1^\theta(\bm{x}|{\bm{x}_s},{\bm{x}_0}) + (\dot{\sigma}_t - \sigma_t\gamma_t)\, U_{\frac{1}{K}}(\bm{x}) + \gamma_t\delta_{\bm{x}_s}(\bm{x}),\label{eq:bwdvelocity}
\end{equation}
Besides, once the parameterized velocity is learned, one can sample the probability path through the one-step Euler ODE solver, as 
\begin{equation}
    \bm{x}_t \sim \delta_{\bm{x}_s}(\cdot) + v_t^{\theta}(\bm{x}_s, \bm{x}_0) h,
\end{equation}
where $h = t - s$ is the sampling interval, after multi-steps of iteration until $t=1$, the final sample $\bm{x}_1\sim \delta_{\bm{x}_{1-h}}(\cdot) + v_1^{\theta}(\bm{x}_{1-h}, \bm{x}_0)$ is the predicted target variable. Similarly, to bridge the reverse process from $t=1$ to $t=0$, the same training and sampling approaches can be used, where the neural network should approximate $\delta_{\bm{x}_0}(\bm{x})$ with $\bm{x}_1$ and $\bm{x}_s$ ($s>t$) as input, leading to
\begin{equation}
     v^{\theta}_t(\bm{x}|\bm{x}_s, \bm{x}_1) = (\dot{\alpha}_t -\alpha_t\gamma_t) p_0^\theta(\bm{x}|{\bm{x}_s,\bm{x}_1}) + (\dot{\beta}_t -\beta_t\gamma_t)\delta_{\bm{x}_1}(\bm{x}) + (\dot{\sigma}_t - \sigma_t\gamma_t)\, U_{\frac{1}{K}}(\bm{x}) + \gamma_t\delta_{\bm{x}_s}(\bm{x}),
\end{equation}
Algorithm.~\ref{alg:dfbtrain} and \ref{alg:dfbsample} gives the pseudo-code of training and sampling of our proposed \textit{discrete flow bridge}.
\begin{algorithm}[H]
\caption{Discrete Flow Bridge training} \label{alg:dfbtrain}
\begin{algorithmic}[1]
\Require neural network $p^{\theta}_1$, source and target samples $\bm{x}_0, \bm{x}_1 \sim \pi(\bm{x}_0, \bm{x}_1)$, learning rate $\eta$, scheduler coefficient $\alpha_t, \beta_t, \sigma_t$
\Repeat
\State Sample $t \sim U_{[0,1]}$ 
\State Sample $\bm{x}_t \sim p_t(\bm{x}|\bm{x}_0,\bm{x}_1)$ with the scheduler $\alpha_t, \beta_t, \sigma_t$ according to Eq.~\ref{eq:cpp}
\State $\mathcal{L}(\theta) \leftarrow \mathrm{CE}(p_1^{\theta}(\bm{x}_t,\bm{x}_0), \delta_{\bm{x}_1}(\cdot) )$
\State $\theta \leftarrow \theta - \eta \nabla_\theta \mathcal{L}(\theta)$
\Until{convergence}
\end{algorithmic}
\end{algorithm}

\begin{algorithm}[H]
\caption{Discrete Flow Bridge sampling}\label{alg:dfbsample}
\begin{algorithmic}[1]
\Require parameterized target distribution $p^\theta_1$, sample $x_0 \sim p_0$, step size $h = \frac{1}{n}$
\For{$t = 0, h, 2h, \dots, 1 - h$}
    \State Calculate parameterized velocity $v^{\theta}_t(\bm{x}_t,\bm{x}_0)$ through Eq.~\ref{eq:fwdvelocity}
    \State $\bm{x}_{t+h} \sim \delta_{\bm{x}_t}(\cdot) + v^{\theta}_t(\bm{x}_t,\bm{x}_0)h$
\EndFor
\State \Return $\bm{x}_1$
\end{algorithmic}
\end{algorithm}
\subsection{Multitask Learning for Reaction Prediction}\label{sec:learning}
\subsubsection{Bridging Reactants and Products}
In our formulation of the reaction prediction problem, four discrete variables from the source graph are considered to be translated to the target graphs. In detail, we simply set the scheduler of changing rates as
\begin{equation}
\begin{aligned}
    \alpha_t &= \left(1 - \sigma\sqrt{t(1-t)}\right) \cdot (1-t);\\
    \beta_t &= \left(1 - \sigma\sqrt{t(1-t)}\right) \cdot t;\\
    \sigma_t &= \sigma\sqrt{t(1-t)}, \label{eq:scheduler}
\end{aligned}
\end{equation}
where $\sigma \in [0,2]$.
By setting the inclusion of 72 types of elements and one dummy type ($M=73$ shown in Appendix.~\ref{app:preprocess}), the four discrete variables follow the conditional paths, which read
\begin{equation}
\begin{aligned}
    p_t(a | a_0, a_1) &= \alpha_t\, \delta_{a_0}(a) + \beta_t\, \delta_{a_1}(a) + \sigma_t\, U_{\frac{1}{73}}(a), \\
    p_t(b | b_0, b_1) &= \alpha_t\, \delta_{b_0}(b) + \beta_t\, \delta_{b_1}(b) + \sigma_t\, U_{\frac{1}{2}}(b), \\
    p_t(c | b_0, b_1) &= \alpha_t\, \delta_{c_0}(c) + \beta_t\, \delta_{c_1}(c) + \sigma_t\, U_{\frac{1}{13}}(c), \\
    p_t(r | r_0, r_1) &= \alpha_t\, \delta_{r_0}(r) + \beta_t\, \delta_{r_1}(r) + \sigma_t\, U_{\frac{1}{4}}(r). \\
\end{aligned}
\end{equation}
In this way, we can firstly sample $\mathcal{G}_t \sim p_t(\mathcal{G}|\mathcal{G}_0, \mathcal{G}_1)$, as 
\begin{equation}
    \begin{aligned}
        \mathcal{G}_t = (\mathcal{V}_t, \mathcal{E}_t) =&\left( (\{a^{(i)}_t\}, \{b^{(i)}_t\}, \{c^{(i)}_t\}) ,\{r^{(i,j)}_t\} \right)_{1\leq i \leq N, 1\leq j\leq N}\\
        \text{where } &a_t^{(i)} \sim  p_t(a | a^{(i)}_0, a^{(i)}_1),  b_t^{(i)} \sim  p_t(b | b^{(i)}_0, b^{(i)}_1), \\
        &c_t^{(i)} \sim  p_t(c | c^{(i)}_0, c^{(i)}_1),  r_t^{(i,j)} \sim  p_t(r | r^{(i,j)}_0, r^{(i,j)}_1). \label{eq:molsample}
    \end{aligned}
\end{equation}
Then, when translating reactants to products as a forward prediction task, to unify the notation, we write the product graph as $\mathcal{G}_+ =\mathcal{G}_1$ and the reactant graph as $\mathcal{G}_- =\mathcal{G}_0$, a neural network $\varphi^{\theta}(\cdot)$ taking $(\mathcal{G}_-, \mathcal{G}_t)$ as input will be used to approximate $\delta_{\bm{x}_1}(\cdot)$ with CE loss:
\begin{equation}
\begin{aligned}
    \mathcal{L}_+(\theta) = \mathbb{E}_{\substack{
(\mathcal{G}_-, \mathcal{G}_+) \sim \pi(\mathcal{G}_-, \mathcal{G}_+) \\
\mathcal{G}_t \sim p_t(\mathcal{G} \mid \mathcal{G}_-, \mathcal{G}_+)
} } \mathrm{CE}\left(\varphi^{\theta}(\mathcal{G}_t,\mathcal{G}_-), \delta_{\mathcal{G}_+}(\cdot) \right). \label{eq:lossfwd}
\end{aligned}
\end{equation}
where $\mathrm{CE}\left(\varphi^{\theta}(\mathcal{G}_t,\mathcal{G}_-), \delta_{\mathcal{G}_+}(\cdot) \right)$ is computed by aggregating the contributions from all elements in the set $\{a_t^{(i)}, b_t^{(i)}, c_t^{(i)}, r_t^{(i,j)}\}$. For iterative generating the product graph, it samples the conditional probability path through
\begin{equation}
    \mathcal{G}_{t+h} \sim \delta_{\mathcal{G}_t}(\cdot) + v_{t+h}^{\theta}(\mathcal{G}_t, \mathcal{G}_-) h,
\end{equation}
where $v_{t+h}^{\theta}(\mathcal{G}_t, \mathcal{G}_-)$ can be computed with $\varphi^{\theta}(\mathcal{G}_t,\mathcal{G}_-)$  according to Eq.~\ref{eq:fwdvelocity}. 

Similarly, for reactant prediction as a retrosynthesis task, the source sample should be set as $\mathcal{G}_-$ instead, and the neural network $\phi^\theta(\cdot)$ approximates  $\delta_{\bm{x}_1}(\cdot)$ with
\begin{equation}
\begin{aligned}
    \mathcal{L}_-(\theta) = \mathbb{E}_{\substack{
(\mathcal{G}_-, \mathcal{G}_+) \sim \pi(\mathcal{G}_-, \mathcal{G}_+) \\
\mathcal{G}_t \sim p_t(\mathcal{G} \mid \mathcal{G}_-, \mathcal{G}_+)
} } \mathrm{CE}\left(\varphi^{\theta}(\mathcal{G}_t,\mathcal{G}_+), \delta_{\mathcal{G}_-}(\cdot) \right), \label{eq:lossbwd}
\end{aligned}
\end{equation}
The sampling process can be obtained by 
\begin{equation}
    \mathcal{G}_{t-h} \sim \delta_{\mathcal{G}_t}(\cdot) - v_{t-h}^{\theta}(\mathcal{G}_t, \mathcal{G}_+) h,
\end{equation}
where $v_{t-h}^{\theta}(\mathcal{G}_t, \mathcal{G}_+)$ can be computed with $\varphi^{\theta}(\mathcal{G}_t,\mathcal{G}_+)$ according to Eq.~\ref{eq:bwdvelocity}.
\subsubsection{Multitask Learning}
To enable our model $\varphi^{\theta}$ to perform multiple tasks—namely, translating from the reactant graph $\mathcal{G}_-$ to the product graph $\mathcal{G}_+$ for forward reaction prediction, and inversely translating the product back to the reactant for retrosynthesis, we introduce an additional task token $\mathrm{x}_{\mathrm{tsk}}$. Similar to a prompt token in large language models~\cite{Raffel2020T5,Irwin2022Chemformer,Jia2022VPT,Zhou2021CoOp,Ross2021MoLFormer,Shen2022MVLPT,Ramesh2021DALL-E,Radford2021CLIP}, this guides the neural network $\varphi^\theta$ to carry out either the forward or retro prediction accordingly. Specifically, $\mathrm{x}_{\mathrm{tsk}}=0$ if the task is product prediction, or $1$ if it is retrosynthesis.

Therefore, $\varphi^\theta$ takes three variables as inputs, $\mathcal{G}_t$, $\mathcal{G}_\mp$ and $\mathrm{x}_\mathrm{tsk}$.  Concretely, we design $\varphi^\theta$ as a modular architecture consisting of three components shown in Fig.~\ref{fig:workflow}.e: \textit{MolEncoder}, which encodes input molecular graphs; \textit{MergeEncoder}, which integrates task tokens and molecular representations; and \textit{MolDecoder}, which generates attributes in target molecular graph.
\paragraph{MolEncoder}
In \textit{MolEncoder}, atom variables, including atom type, binary aromatic flag, and formal charge, are first mapped into a high-dimensional space through separate linear layers. The resulting embeddings are summed to form the atom representation $\bm{\mathrm{{z}}}_\mathrm{in} \in \mathbb{R}^{N \times D}$. 
As shown in Fig.~\ref{fig:workflow}.d, a message-passing layer regulated by the adjacency matrix $\bm{R}$ integrates bond-related information and produces
\begin{equation}
    \bm{\mathrm{{z}}}_\mathrm{msg} = \bm{R} \bm{\mathrm{{z}}}_\mathrm{in}.
\end{equation}
This atom-level representation is further processed by a stack of transformer layers with self-attention mechanisms, yielding the encoded output:
\begin{equation}
    \bm{\mathrm{{z}}}_\mathrm{enc} = \mathrm{Transformers}(\bm{\mathrm{{z}}}_\mathrm{msg}).
\end{equation}
Since both the intermediate graph $\mathcal{G}_t$ and the source graph $\mathcal{G}_{\mp}$ need to be encoded, we employ two separate instances of the \textit{MolEncoder}. After extracting their respective atom-level representations $\bm{\mathrm{{z}}}_{\mathrm{enc},t}$ and $\bm{\mathrm{{z}}}_{\mathrm{enc},\mp}$, we sum them to obtain the final joint representation:
\[
    \bm{\mathrm{{z}}}'_\mathrm{enc} = \bm{\mathrm{{z}}}_{\mathrm{enc},t} + \bm{\mathrm{{z}}}_{\mathrm{enc},\mp}.
\]
\paragraph{MergeEncoder}
To merge the task token information into the molecular representation, we firstly lift $\mathrm{x}_\mathrm{tsk}$ into $D$-dimensional latent space as $\bm{\mathrm{{z}}}_\mathrm{tsk} \in \mathbb{R}^{1\times D}$, and use multi-layers of cross-attention Transformers like \cite{zhang2023adding,raffel2020exploring,tamkin2021language,li2021prefix}, with the $\bm{\mathrm{{z}}}_\mathrm{tsk}$ as query and the $\bm{\mathrm{{z}}}'_\mathrm{enc}$ as key and value:
\begin{equation}
\begin{aligned}
    \bm{\mathrm{{z}}}'_\mathrm{merge} &= \mathrm{CrossAttnTransformers}(\bm{\mathrm{{z}}}_\mathrm{tsk}, \bm{\mathrm{{z}}}'_\mathrm{enc})
\end{aligned}
\end{equation}
where a single layer 
\begin{equation}
    \begin{aligned}
        \mathrm{CrossAttnTransformer}(\bm{\mathrm{{z}}}_\mathrm{tsk}, \bm{\mathrm{{z}}}'_\mathrm{enc}) = \mathrm{FFN}\left(\mathrm{LN}(\mathrm{CrossAttn}(\bm{\mathrm{{z}}}_\mathrm{tsk}, \bm{\mathrm{{z}}}'_\mathrm{enc})) + \bm{\mathrm{{z}}}'_\mathrm{enc}\right),
    \end{aligned}
\end{equation}
in which $\mathrm{LN}$ is the layer normalization, and $\mathrm{FFN}$ is the feed-forward layer, leading to  $\bm{\mathrm{{z}}}'_\mathrm{merge} \in \mathbb{R}^{N\times D}$.
\paragraph{MolDecoder}
As shown in Fig.~\ref{fig:workflow}.f, the architecture of \textit{MolDecoder} consists of multiple Transformer layers with self-attention mechanisms, which transform the merged representation $\bm{\mathrm{z}}'_\mathrm{merge}$ into the final atom-level molecular representation $\bm{\mathrm{z}}_\mathrm{dec}$. 
Three separate linear heads of $\{\mathrm{Linear}_{\mathrm{a}}$, $\mathrm{Linear}_{\mathrm{b}}\}$, $\mathrm{Linear}_{\mathrm{c}}$ are applied to $\bm{\mathrm{z}}_\mathrm{dec}$ to produce the probabilistic logits for atom type, binary aromaticity, and formal charge, respectively.
\begin{equation}
\begin{aligned}
    \mathrm{logit}_\mathrm{a} &= \mathrm{Linear}_{\mathrm{a}}(\bm{\mathrm{{z}}}_\mathrm{dec}) \in \mathbb{R}^{N\times 73},\\
\mathrm{logit}_\mathrm{b} &=  \mathrm{Linear}_{\mathrm{b}}(\bm{\mathrm{{z}}}_\mathrm{dec}) \in \mathbb{R}^{N\times 2},\\
\mathrm{logit}_\mathrm{c} &= \mathrm{Linear}_{\mathrm{c}}(\bm{\mathrm{{z}}}_\mathrm{dec}) \in \mathbb{R}^{N\times 13},
\end{aligned}
\end{equation}
Besides, a self-attention matrix is employed as the logits of adjacency, which reads 
\begin{equation}
    \mathrm{logit}_\mathrm{r} = (\mathrm{Linear}_{\mathrm{qry}}(\bm{\mathrm{{z}}}_\mathrm{dec}))^{\mathsf{T}}\mathrm{Linear}_{\mathrm{key}}(\bm{\mathrm{{z}}}_\mathrm{dec}) \in \mathbb{R}^{N\times N\times 4}.
\end{equation}
\paragraph{Training and Sampling}
The logits $\{\mathrm{logit}_\mathrm{a}, \mathrm{logit}_\mathrm{b}, \mathrm{logit}_\mathrm{c}, \mathrm{logit}_\mathrm{r}\}$ produced by the \textit{MolDecoder} are used to compute the cross-entropy loss as defined in Eq.~\ref{eq:lossfwd} and Eq.~\ref{eq:lossbwd}, corresponding to the forward and retro prediction objectives, respectively. 
The final loss is calculated by averaging over the atom-node dimension ($N$) for atom-level variables, and over the bond adjacency matrix dimension ($N \times N$) for bond-level variables. The total loss is obtained by summing the outputs of the four decoding heads.
To enable the model to learn the bidirectional mapping between $\mathcal{G}_-$ and $\mathcal{G}_+$, we randomly sample either a `product prediction' or a `reactant prediction' task at each training iteration. The overall training procedure for our model is summarized in the following Algorithm.~\ref{alg:synbtrain}:
\begin{algorithm}[H]
\caption{SynBridge training for multi-task}\label{alg:synbtrain}
\begin{algorithmic}[1]
\Require Neural network $\varphi^\theta$, source and target samples $\mathcal{G}_-, \mathcal{G}_+ \sim \pi(\mathcal{G}_-, \mathcal{G}_+)$, learning rate $\eta$, scheduler coefficients $\alpha_t, \beta_t, \sigma_t$ (see Eq.~\ref{eq:scheduler})
\Repeat
    \State Sample $t \sim \mathcal{U}[0, 1]$
    \State Sample task token $\mathrm{x}_\mathrm{tsk} \in \{0, 1\}$ \Comment{Uniformly sample task token}
    \State Sample intermediate graph $\mathcal{G}_t \sim p_t(\mathcal{G} | \mathcal{G}_-, \mathcal{G}_+)$ with the scheduler in Eq.~\ref{eq:scheduler}, based on Eq.~\ref{eq:molsample}
    \If{$\mathrm{x}_\mathrm{tsk} = 0$} \Comment{Forward product prediction}
        \State $\mathcal{L}(\theta) \gets \mathrm{CE}\left(\phi^{\theta}(\mathcal{G}_t, \mathcal{G}_-,\mathrm{x}_\mathrm{tsk}), \delta_{\mathcal{G}_+}(\cdot)\right)$
    \Else \Comment{Retro reactant prediction}
        \State $\mathcal{L}(\theta) \gets \mathrm{CE}\left(\phi^{\theta}(\mathcal{G}_t, \mathcal{G}_+,\mathrm{x}_\mathrm{tsk}), \delta_{\mathcal{G}_-}(\cdot)\right)$
    \EndIf
    \State $\theta \gets \theta - \eta \nabla_\theta \mathcal{L}(\theta)$
\Until{convergence}
\end{algorithmic}
\end{algorithm}
After training, the forward and backward conditional velocities can be computed using Eq.~\ref{eq:fwdvelocity} or Eq.~\ref{eq:bwdvelocity}, respectively. This enables sampling to be performed either in the forward direction (from $t = 0$ to $t = 1$), performing generation from reactants to products, or in the reverse direction (from $t = 1$ to $t = 0$), representing generation from products to reactants.

\section{Experiment}
\subsection{Implementation Details and Setup}
SynBridge is implemented in \texttt{Python 3.9}, with RDKit (\texttt{version 2024.9.6}) serving as the primary toolkit for molecule parsing and reconstruction—specifically, for converting SMILES strings into atom- and bond-level variables, and for recovering valid SMILES from predicted discrete variables.
As the core of SynBridge framework, the discrete flow bridge module is implemented using PyTorch (\texttt{v2.6.0+cu124}). The transformer-based modules leverage PyTorch’s integrated \texttt{TransformerEncoderLayer} with fused attention kernels for enhanced computational efficiency.
SynBridge's training was performed using a cluster of 4 NVIDIA A100 GPUs (each with 81920 MiB memory), and optimization was carried out using the AdamW optimizer with a learning rate of 1e-4. In all the trials, SynBridge is trained for {400k} iterations with {5k} warming-up iterations, as the diffusion-based models typically require longer convergence training steps due to their stochastic sampling nature.
In the molecular representation, we include a vocabulary of 72 unique atomic types (shown in Appendix.~\ref{app:preprocess}), 4 bond types (including `\texttt{nonbonded}'), atomic formal charges ranging from -6 to {+6}, and binary aromaticity ({\{0,1\}}), each treated as separate discrete variables for atom- and bond-level prediction. The latent dimension $D$ is uniformly set to 256 across all layers, while the number of transformer layers varies, as specified in Table~\ref{tab:hyperparam1},~\ref{tab:hyperparam2} and ~\ref{tab:hyperparam3} in Appendix~\ref{app:expdetail}. In addition, we find that in the sampling, $\sigma_t$ is set to $0$ can bring more deterministic results with higher prediction accuracy, so we simplify the sampling algorithm in practice.
The source code and trained models are publicly available at \texttt{\url{https://github.com/EDAPINENUT/synbridge}} to facilitate future research.

We select several state-of-the-art models as baselines for comprehensive comparison, including:
\textcircled{1}NERF\cite{bi2021nerf} – a one-shot, graph-based variational autoencoder for forward product prediction;
\textcircled{2}MEGAN\cite{sacha2021megan} – an auto-regressive, graph-based model that supports both forward and retrosynthetic prediction;
\textcircled{3}RetroBridge\cite{igashov2024retrobridge} – a one-shot graph-based diffusion bridge model tailored for retrosynthesis;
\textcircled{4}T5Chem\cite{kim2021t5chem} – an auto-regressive, sequence-based language model capable of multitask prediction.
In addition, to more rigorously assess the performance of SynBridge, we introduce two internal baseline models:
\textcircled{5}G2G-Former – a graph-to-graph transformer architecture that shares the same network backbone as SynBridge but omits the discrete flow bridge module. It functions as a one-step deterministic variant of SynBridge, performing direct mapping from the input graph to the output graph in a single pass;
\textcircled{6}S2S-Former – a conventional sequence-to-sequence transformer that takes SMILES as input and generates SMILES as output, following the standard paradigm in SMILES-based reaction prediction.
For the implementation of \textcircled{5} and \textcircled{6}, we apply a softmax function with a temperature $\tau=1$ to the classification logits during the final sampling stage, converting them into probabilities. We then perform multinomial sampling based on these probabilities to generate the product or reactant graphs.
\begin{table}[ht]
\centering
\caption{Discreption of datasets used for our experiments}\label{tab:dataset_splits}

\begin{tabular}{lrrrrc}
\toprule
\textbf{Dataset} & \textbf{\#Train} & \textbf{\#Valid} & \textbf{\#Test} & \textbf{\#Total} & \textbf{Task} \\
\midrule
USPTO-50K~\cite{dai2019retrosynthesis} & 40,000 & 5,000 & 5,000 & 50,000 & Forward / Retrosynthesis \\
USPTO-MIT$^{*}$~\cite{jin2017predicting} & 409,000 & 30,000 & 40,000 & 479,000 & Forward / Retrosynthesis \\
Pistachio$^{*}$~\cite{schwaller2021rxnmapper} & 2,508,278 & 2,000 & 300,000 & 2,810,278 & Forward / Retrosynthesis \\
\bottomrule
\end{tabular}
\begin{tablenotes}
\small
\item[$*$] Contains reactions involving ionic species and atomic charge changes.
\end{tablenotes}
\end{table}
\subsection{Dataset}

We consider three chemical reaction datasets of increasing scale: \textbf{USPTO-50K}, \textbf{USPTO-MIT}, and \textbf{Pistachio}.

\textbf{USPTO-50K} is commonly used as a benchmark for single-step retrosynthesis. This dataset is a curated subset of Lowe’s patent database, comprising 50,000 reactions categorized into 10 general reaction classes. We adopt the standard data split as proposed by \cite{dai2019retrosynthesis}, leading to 40,000, 5000, and 5000 reactions used for training, validation, and testing, respectively.
Although our model is invariant to the ordering of graph nodes, we use the version of USPTO-50K with canonical SMILES, as released by \cite{somnath2021learning}, to address a dataset artifact they identified: in nearly 75\% of reactions, the product atom labeled with atom-mapping number 1 is directly involved in the transformation. Due to its relatively small size and the simplicity of the chemical reactions it contains, with no changes in atomic charges and no ionic species present in either reactants or products, we consider this dataset the simplest among our benchmarks. It serves as a baseline for evaluating model performance on both forward product prediction and retrosynthesis tasks.

\textbf{USPTO-MIT}  is a widely used benchmark dataset, which comprises approximately 480,000 chemical reactions curated from the original USPTO data compiled by Lowe. The dataset was constructed by \cite{jin2017predicting} through a rigorous cleaning process that removed duplicates and erroneous entries, and retained only reactions with well-defined, contiguous reaction centers. The dataset is split into 409,000 reactions for training, 30,000 for validation, and 40,000 for testing, and has become a standard benchmark for evaluating the performance of reaction prediction models under both forward and retrosynthesis scenarios. USPTO-MIT includes information on reactants, reaction conditions (reagents), and products. In this work, we focus solely on the prediction of reactants to products and vice versa. Compared to USPTO-50K, the USPTO-MIT dataset poses a greater challenge as it includes reactions involving ionic species and changes in atomic charges.

\textbf{Pistachio} is a large-scale reaction dataset derived from the Pistachio database developed by NextMove Software. It contains 16,678,201 chemical reactions extracted from patent literature, making it one of the most comprehensive publicly accessible reaction corpora. Compared to USPTO-50K and USPTO-MIT, the Pistachio dataset is significantly larger in scale and exhibits a much broader diversity in reaction types, reactant complexity, and chemical transformations. While Pistachio includes various metadata such as reaction conditions and textual annotations, we focus on the core reaction information—reactants and products—for both forward and retrosynthesis modeling. The dataset naturally includes challenging phenomena such as ionic species and thousands of reaction types, enables the training of large models, and facilitates the evaluation of generalization capabilities in more diverse chemical environments. For SMILES preprocessing, we first validate the chemical validity of each reactant and product using RDKit. We then apply atom mapping using \texttt{RXNMapper}~\cite{schwaller2021rxnmapper}, retaining only reactions with a confidence score greater than 0.8. Additionally, we discard any reactions in which atoms in the product cannot be matched to atom-mapped atoms in the reactants. To ensure compatibility with our model’s capacity, we further filter out reactions in which the number of atoms in the reactants exceeds 80. After preprocessing, we obtain a final dataset consisting of 478,192 reactions, which we randomly split into 408,192 for training, 30,000 for validation, and 40,000 for testing, respectively.

\subsection{Result and Discussion}
\subsubsection{Single Task Evaluation}
\begin{figure}[ht]
\centering

% ---------- Forward ----------
\begin{minipage}{0.495\linewidth}
\centering
\text{\small Forward}\vspace{0.17em}
\resizebox{\linewidth}{!}{
\begin{tabular}{lccc}
\toprule
{Model} & {Top 1 (\%)} & {Top 3 (\%)} & {Top 5 (\%)} \\
\midrule
\rowcolor{gray!10} MEGAN & 88.9 & 90.8 & 93.0 \\
NeRF & 94.6 & 96.1 & \textbf{97.9} \\
\rowcolor{gray!10} RetroBridge & -- & -- & -- \\
\multirow{2}{*}{\shortstack[l]{RetroBridge \\ (\#atom-prior)}} & \multirow{2}{*}{--} & \multirow{2}{*}{--} & \multirow{2}{*}{--} \\
 & & & \\
\rowcolor{gray!10}G2G-Former & 93.0 & 93.9 & 94.0 \\
 S2S-Former & 95.0 & 95.2 & 95.8 \\
\rowcolor{gray!10} SynBridge & \textbf{95.9} & \textbf{96.2} & {96.5} \\
\bottomrule
\end{tabular}
}
\end{minipage}
% ---------- Retrosynthesis ----------
\begin{minipage}{0.495\linewidth}
\centering
\text{\small Retrosynthesis}
\resizebox{\linewidth}{!}{
\begin{tabular}{lccc}
\toprule
{Model} & {Top 1 (\%)} & {Top 3 (\%)} & {Top 5 (\%)} \\
\midrule
\rowcolor{gray!10} MEGAN & 48.8 & 63.5 & 71.6 \\
NeRF & -- & -- & -- \\
\rowcolor{gray!10} RetroBridge & 50.9 & 74.1 & 80.8 \\
\multirow{2}{*}{\shortstack[l]{RetroBridge \\ (\#atom-prior)}} & \multirow{2}{*}{61.6} & \multirow{2}{*}{82.5} & \multirow{2}{*}{85.4} \\
 & & & \\
\rowcolor{gray!10}G2G-Former & 53.8 & 77.0 & 79.2 \\
 S2S-Former & 45.4 & 58.7 & 66.4 \\
\rowcolor{gray!10}SynBridge & \textbf{79.4} & \textbf{84.4} & \textbf{85.7} \\
\bottomrule
\end{tabular}
}
\end{minipage}

\captionof{table}{Model performance comparison (Top-k accuracy in \%) on USPTO-50K for both forward prediction and retrosynthesis tasks.}
\label{tab:uspto_50k_comparison}
\end{figure}

\begin{figure}[ht]
\centering

\begin{minipage}{0.495\linewidth}
\centering 
\text{\small Forward }  
\vspace{-0.15em}
\resizebox{\linewidth}{!}{
\begin{tabular}{lccc}
\toprule
{Model} & {Top 1 (\%)} & {Top 3 (\%)} & {Top 5 (\%)} \\
\midrule
\rowcolor{gray!10}MEGAN & 86.5 & 90.5 & 91.9 \\
NeRF & 87.1 & 88.8 & 89.1 \\
\rowcolor{gray!10}G2G-Former & 86.4 & 89.2 & 90.0 \\
S2S-Former & 86.6 & \textbf{90.6} & 92.2 \\
\rowcolor{gray!10}SynBridge & \textbf{88.4} & {90.1} & \textbf{92.4} \\
\bottomrule
\end{tabular}
}
\end{minipage}\hspace{0.1em}
\begin{minipage}{0.495\linewidth}
\centering
\text{ \small Retrosynthesis}
\resizebox{\linewidth}{!}{
\begin{tabular}{lccc}
\toprule
{Model} & {Top 1 (\%)} & {Top 3 (\%)} & {Top 5 (\%)} \\
\midrule
\rowcolor{gray!10}MEGAN & 7.3 & 8.4 & 9.0 \\
NeRF & -- & -- & -- \\
\rowcolor{gray!10}G2G-Former & 25.6 & 32.8 & 34.6 \\
S2S-Former & 10.0 & 17.6 & 20.0 \\
\rowcolor{gray!10}SynBridge & \textbf{37.8} & \textbf{40.1} & \textbf{41.5} \\
\bottomrule
\end{tabular}
}
\end{minipage}

\captionof{table}{Model performance comparison (Top-k accuracy in \%) on USPTO-MIT for both forward prediction and retrosynthesis tasks.}
\label{tab:uspto_mit_comparison}
\end{figure}

\begin{figure}[ht]
\centering

% ---------- Forward ----------
\begin{minipage}{0.495\linewidth}
\centering 
\text{\small Forward }  
\vspace{-0.15em}
\resizebox{\linewidth}{!}{
\begin{tabular}{lccc}
\toprule
{Model} & {Top 1 (\%)} & {Top 3 (\%)} & {Top 5 (\%)} \\
\midrule
\rowcolor{gray!10}MEGAN & 86.9 & 93.1 & 96.9 \\
NeRF & 92.2 & 94.6 & 95.9 \\
\rowcolor{gray!10}G2G-Former & 90.4 & 93.0 & 93.6 \\
S2S-Former & 91.0 & 94.2 & 96.4 \\
\rowcolor{gray!10}SynBridge & \textbf{94.8} & \textbf{97.4} & \textbf{98.0} \\
\bottomrule
\end{tabular}
}
\end{minipage}\hspace{0.1em}
% ---------- Retrosynthesis ----------
\begin{minipage}{0.495\linewidth}
\centering
\text{\small Retrosynthesis}
\resizebox{\linewidth}{!}{
\begin{tabular}{lccc}
\toprule
{Model} & {Top 1 (\%)} & {Top 3 (\%)} & {Top 5 (\%)} \\
\midrule
\rowcolor{gray!10}MEGAN & 33.3 & 42.7 & 45.4 \\
NeRF & -- & -- & -- \\
\rowcolor{gray!10}G2G-Former & 55.6 & 66.5 & 66.7 \\
S2S-Former & 37.8 & 43.1 & 44.7 \\
\rowcolor{gray!10}SynBridge & \textbf{66.1} & \textbf{75.0} & \textbf{79.3} \\
\bottomrule
\end{tabular}
}
\end{minipage}

\captionof{table}{Model performance comparison (Top-k accuracy in \%) on Pistachio for both forward prediction and retrosynthesis tasks.}
\label{tab:pistachio_comparison}
\end{figure}

\begin{table}[]
\begin{tabular}{c c l c c c}
\toprule
Dataset & Task     & Model     & Top 1(\%) & Top 3(\%) & Top 5(\%) \\
\midrule
\multirow{4}{*}{USPTO-50K} &
\multirow{2}{*}{Forward} 
    & T5Chem    & 93.6 & 94.1 & 94.5 \\
&   & SynBridge & \textbf{94.4} & \textbf{95.9} & \textbf{96.2} \\
\cmidrule(lr){2-6}
& \multirow{2}{*}{Retro}   
    & T5Chem    & 44.7 & 56.6 & 71.5 \\
&   & SynBridge & \textbf{81.9} & \textbf{85.8} & \textbf{86.7} \\
\midrule
\multirow{4}{*}{USPTO-MIT} &
\multirow{2}{*}{Forward} 
    & T5Chem    & \textbf{87.8} & \textbf{92.1} & 93.9 \\
&   & SynBridge & 83.7 & 90.9 & \textbf{94.8} \\
\cmidrule(lr){2-6}
& \multirow{2}{*}{Retro}   
    & T5Chem    & 11.6 & 19.2 & 23.1 \\
&   & SynBridge & \textbf{39.8} & \textbf{42.9} & \textbf{44.2} \\
\midrule
\multirow{4}{*}{Pistachio} &
\multirow{2}{*}{Forward} 
    & T5Chem    & 93.1 & 97.2 & 98.1 \\
&   & SynBridge & \textbf{95.4} & \textbf{98.1} & \textbf{98.6} \\
\cmidrule(lr){2-6}
& \multirow{2}{*}{Retro}   
    & T5Chem    & 67.4 & 80.3 & 82.9 \\
&   & SynBridge & \textbf{71.5} & \textbf{81.9} & \textbf{84.3} \\
\bottomrule
\end{tabular}

\captionof{table}{Model performance comparison (Top-k accuracy in \%) on multi-task training.}\label{tab:mtcomparison}
\end{table}
We conducted separate experimental evaluations on the three datasets with the two tasks mentioned above. Our primary metric is Top-k accuracy, which measures the proportion of cases in which the correct reactants/products appear among the top-k predictions generated by the model.
Our method consistently achieved state-of-the-art performance across nearly all experiments. In particular, on the Pistachio dataset—recognized for its high chemical complexity—we significantly outperformed the second-best method in both the forward and retrosynthesis tasks, with absolute improvements of 2.6\% and 10.5\%, respectively.

In the overall task evaluation, we observed that the performance improvement of our model on the retrosynthesis task was considerably larger compared to baseline models. We attribute this to the fact that SynBridge is a bidirectional model, which benefits from a strong prior: during retrosynthesis, the total number of atoms is given, and the model directly predicts the atomic types. To exploit this prior, we developed G2G-Former, an end-to-end graph-based model that also operates under the assumption of a known atom count.

Furthermore, we adapt a flow-based generative model for retrosynthesis, RetroBridge, by conditioning it on the number of atoms denoted RetroBridge (\#atom-prior). Even with this prior, RetroBridge exhibited a clear performance gap compared to SynBridge on the USPTO-50k dataset (61.6\% vs. 79.4\% in accuracy). Notably, RetroBridge does not model atomic charges, making it unsuitable for generalization to the USPTO-MIT and Pistachio datasets.
Beyond these primary conclusions, we also observe the following:
	\begin{itemize}
	    \item Although the Pistachio dataset is generally considered more chemically complex, its large scale provides a scaling-up effect that enables the model to learn more accurate predictions for complex reactions.
        \item While natural language-based models (e.g., S2S-Former) lack interpretability, they achieve high prediction accuracy. This can be attributed to the inherent compatibility between SMILES as a chemical sequence language and the autoregressive nature of sequence-to-sequence models.
	\end{itemize}
\subsubsection{Multi-Task Evaluation}
Next, we proceed to evaluate our approach on bidirectional prediction through multi-task training, as described in Algorithm.~\ref{alg:synbtrain}. For performance comparison, we adopt T5Chem, the current state-of-the-art multi-task chemical language-based auto-regressive model. From the Table.~\ref{tab:mtcomparison}, we conclude the following:
	\begin{itemize}
	    \item Under the multi-task training setting, SynBridge generally outperforms T5Chem in terms of predictive performance, with the exception of the USPTO-MIT forward product prediction, where it slightly underperforms. Notably, the improvement is most significant on the retrosynthesis task. We attribute this, in part, to the fact that SynBridge operates with the atom number as prior knowledge, whereas T5Chem performs autoregressive SMILES generation without access to the true atom count.
        \item 	Compared to the single-task training setting, SynBridge exhibits a slight performance drop in forward prediction, but improvements in retrosynthesis accuracy. For instance, on USPTO-50K, the forward Top-1 accuracy drops by 1.5\%, whereas the retrosynthesis Top-1 accuracy increases by 2.5\%. This phenomenon can be explained by the nature of multi-task learning as a form of multi-objective optimization: during training, the model tends to favor more challenging tasks, and a marginal compromise in easier tasks can serve as a trade-off that benefits performance on the more difficult ones.
	\end{itemize}

\subsection{Ablation Study} \label{sec:ablation}
To figure out what modules contribute to the performance, we here conduct an ablation study on the forward prediction task of USPTO-MIT dataset. In detail, we consider (1) SynBridge with/without the initial graph state $\mathcal{G}_{\pm}$ as input; (2) SynBridge with different uniform noise level $\sigma_t = \sigma \sqrt{t(1-t)}$ where $\sigma \in [0,2]$; (3) Sampling process with different step size $h=\frac{1}{n}$, where $n$ is the number of sampling steps. The reported setting in Table.~\ref{tab:uspto_mit_comparison} is $\mathcal{G}_+$ and $\mathcal{G}_t$ are both used as input, and $\sigma=1$, number of sampling step $n=20$.
\begin{table}[]
\begin{tabular}{clccc}
\toprule
    \multicolumn{2}{c}{Setting}                  & Top 1(\%) & Top 2(\%) & Top 3(\%) \\
\midrule
\multicolumn{2}{c}{without input $\mathcal{G}_+$}           & 66.4      & 70.8      & 77.6      \\
\cmidrule(lr){1-5}
\multirow{3}{*}{noise levels}      & $\sigma=0.0$   & 77.5      & 83.7      & 85.1      \\
                                   & $\sigma=0.5$ & 82.1      & 85.0      & 85.2      \\
                                   & $\sigma=2.0$ & 85.7      & 86.9      & 88.3      \\
\cmidrule(lr){1-5}
\multirow{3}{*}{\#sampling steps} & n=1          & 86.2      & 88.7      & 89.4      \\
                                   & n=10         & 87.9      & 90.2      & 91.9      \\
                                   & n=100        & 89.5      & 90.9      & 93.6      \\
\cmidrule(lr){1-5}
\multicolumn{2}{c}{with $\mathcal{G}_+, \sigma=1,n=20$}              & 88.4      & 90.1      & 92.4     \\
\bottomrule
\end{tabular}
\caption{Ablation study on different parameters} \label{tab:ablation}
\end{table}
From Table~\ref{tab:ablation}, we can draw the following conclusions:

\begin{enumerate}
    \item[(1)] The use of $\mathcal{G}_{\pm}$ as the initial input state is crucial. This is because the initial state carries significant chemical meaning: the bonds between atoms in the reactants/products serve as structural constraints that directly influence the message-passing process in the graph neural networks. If only the intermediate noisy state is provided as input, the model may lack sufficient informative context, making it difficult to generate accurate predictions.
    \item[(2)] Different uniform noise levels have a significant impact on the results. When the noise level is set to $\sigma = 0.0$, there is no uniform noise introduced during the bridging process, which leads to overconfidence in the generation process. Specifically, once the model makes incorrect type transitions in the early steps, it tends to focus correction only on those positions where $\mathcal{G}_t$ deviates from $\mathcal{G}_{\pm}$, neglecting broader structural uncertainty. When the noise level is too low, this overconfidence issue persists and remains difficult to resolve. On the other hand, if the noise level is too high, the intermediate state $\mathcal{G}_t$ becomes too noisy to provide meaningful guidance for training, ultimately resulting in degraded performance.
    \item[(3)] In the generation process, increasing the number of sampling steps generally leads to improved performance. This is because a higher number of steps allows the model to perform more self-correction operations over the discrete variables, thereby enabling more accurate prediction of the product or reactant as the final state. When $n=1$, the model's performance is comparable to that of G2G-Former, which can be regarded as a one-step end-to-end model. When $n=100$, the performance is similar to that at $n=20$, indicating that the benefit of increasing the number of sampling steps gradually saturates.
\end{enumerate}

\section{Conclusion}
In this work, we present SynBridge, a discrete flow-based bridge model for modeling molecular transformations from reactants to products and vice versa. SynBridge achieves highly accurate generation of reactants and products given the other states of reaction. Through bidirectional modeling on multi-task training, the method supports both forward and retrosynthesis tasks in a unified framework. Extensive experiments on benchmark datasets demonstrate that SynBridge achieves the state-of-the-art performance. Ablation studies confirm the importance of initial graph inputs, uniform noise scheduling, and multi-step sampling for improved accuracy. We believe SynBridge offers a promising direction for product and reactant generation and opens new possibilities for interpretable and chemically consistent reaction modeling.

\section{Data and Software Availability}
On the data side, we provide access to the public datasets USPTO-50K and USPTO-MIT, along with preprocessing scripts for training and evaluation. Due to licensing restrictions, the Pistachio dataset cannot be publicly released.
Our open-source implementation is available at \url{https://github.com/EDAPINENUT/synbridge}, which includes data preprocessing, postprocessing, model training, and evaluation pipelines.
\bibliography{sn-bibliography}
\newpage
\begin{appendices}





\section{Data Preprocess} \label{app:preprocess}
We consider the 72 atomic element types as:

\{ \texttt{H}, \texttt{He}, \texttt{Li}, \texttt{Be}, \texttt{B}, \texttt{C}, \texttt{N}, \texttt{O}, \texttt{F}, \texttt{Na},
\texttt{Mg}, \texttt{Al}, \texttt{Si}, \texttt{P}, \texttt{S}, \texttt{Cl}, \texttt{Ar}, \texttt{K}, \texttt{Ca}, \texttt{Sc},
\texttt{Ti}, \texttt{V}, \texttt{Cr}, \texttt{Mn}, \texttt{Fe}, \texttt{Co}, \texttt{Ni}, \texttt{Cu}, \texttt{Zn}, \texttt{Ga},
\texttt{Ge}, \texttt{As}, \texttt{Se}, \texttt{Br}, \texttt{Rb}, \texttt{Sr}, \texttt{Y}, \texttt{Zr}, \texttt{Mo}, \texttt{Ru},
\texttt{Rh}, \texttt{Pd}, \texttt{Ag}, \texttt{Cd}, \texttt{In}, \texttt{Sn}, \texttt{Sb}, \texttt{Te}, \texttt{I}, \texttt{Xe},
\texttt{Cs}, \texttt{Ba}, \texttt{La}, \texttt{Ce}, \texttt{Pr}, \texttt{Nd}, \texttt{Sm}, \texttt{Eu}, \texttt{Dy}, \texttt{Yb},
\texttt{Hf}, \texttt{Ta}, \texttt{W}, \texttt{Re}, \texttt{Os}, \texttt{Ir}, \texttt{Pt}, \texttt{Au}, \texttt{Hg}, \texttt{Tl},
\texttt{Pb}, \texttt{Bi} \}

If the element type in the reaction is not in these 72 types, we will exclude it from our dataset.
Subsequently, we perform filtering based on atom maps.
For datasets like USPTO-50K and USPTO-MIT that already include atom maps, reactions are filtered out if any atom in the product has an atom map that does not appear in the reactants.
For datasets like Pistachio, which do not contain atom maps, we first generate atom maps using RXNMapper, retain only mappings with a confidence score greater than 0.8, and then apply the same filtering process as above.
We do not consider reaction conditions such as catalysts or reagents, so this information is removed.
\section{Experimental Details} \label{app:expdetail}
\begin{table}[h]
\begin{tabular}{cccc}
\toprule
   Hyper-param               & USPTO-50K & USPTO-MIT & Pistachio \\
\midrule
Learning\_rate    & 1e-4      & 1e-4      & 1e-4      \\
Batch\_size       & 128       & 1024      & 1024      \\
Latent\_dim       & 256       & 256       & 256       \\
\#MolEnc\_layer   & 6         & 6        & 12        \\
\#MergeEnc\_layer & 6        & 6        & 6         \\
\#MolDec\_layer   & 12        & 12        & 24        \\
$\sigma$          & 1         & 1         & 1        \\
\bottomrule
\end{tabular}
\caption{Hyper-parameters for Forward Prediction Training} \label{tab:hyperparam1}
\end{table}
\begin{table}[h]
\begin{tabular}{cccc}
\toprule
   Hyper-param               & USPTO-50K & USPTO-MIT & Pistachio \\
\midrule
Learning\_rate    & 1e-4      & 1e-4      & 1e-4      \\
Batch\_size       & 128       & 1024      & 1024      \\
Latent\_dim       & 256       & 256       & 256       \\
\#MolEnc\_layer   & 6         & 6        & 12        \\
\#MergeEnc\_layer & 6        & 12        & 6         \\
\#MolDec\_layer   & 12        & 12        & 24        \\
$\sigma$          & 1         & 1         & 1        \\
\bottomrule
\end{tabular}
\caption{Hyper-parameters for Retro-synthesis Training} \label{tab:hyperparam2}
\end{table}
\begin{table}[h]
\begin{tabular}{cccc}
\toprule
   Hyper-param               & USPTO-50K & USPTO-MIT & Pistachio \\
\midrule
Learning\_rate    & 1e-4      & 1e-4      & 1e-4      \\
Batch\_size       & 512       & 1024      & 1024      \\
Latent\_dim       & 256       & 256       & 256       \\
\#MolEnc\_layer   & 6         & 12        & 12        \\
\#MergeEnc\_layer & 12        & 12        & 1         \\
\#MolDec\_layer   & 12        & 12        & 24        \\
$\sigma$          & 1         & 1         & 1        \\
\bottomrule
\end{tabular}
\caption{Hyper-parameters for Multi-task Training} \label{tab:hyperparam3}
\end{table}

\end{appendices}

\end{document}